\documentclass[conference]{IEEEtran}
\IEEEoverridecommandlockouts
\usepackage{cite}
\usepackage{amsmath,amssymb,amsfonts}
\usepackage{algorithmic}
\usepackage{graphicx}
\usepackage{textcomp}
\usepackage{xcolor}
\usepackage{graphicx}
\usepackage{booktabs}
\usepackage{comment}
\usepackage{bbm, dsfont}

\def\BibTeX{{\rm B\kern-.05em{\sc i\kern-.025em b}\kern-.08em
    T\kern-.1667em\lower.7ex\hbox{E}\kern-.125emX}}
\begin{document}

\title{Instance-Level Trojan Attacks on Visual Question Answering via Adversarial Learning in Neuron Activation Space}

\author{
    \IEEEauthorblockN{Yuwei Sun\IEEEauthorrefmark{1}\,\IEEEauthorrefmark{2}, Hideya Ochiai\IEEEauthorrefmark{1}, Jun Sakuma\IEEEauthorrefmark{2}\,\IEEEauthorrefmark{3}}
    \IEEEauthorblockA{\IEEEauthorrefmark{1}The University of Tokyo
    \\}
    \IEEEauthorblockA{\IEEEauthorrefmark{2}RIKEN AIP
    \\}
     \IEEEauthorblockA{\IEEEauthorrefmark{3}Tokyo Institute of Technology
    \\}
    {ywsun@g.ecc.u-tokyo.ac.jp, ochiai@elab.ic.i.u-tokyo.ac.jp, sakuma@c.titech.ac.jp}
}
\maketitle

\begin{abstract}
Trojan attacks embed perturbations in input data leading to malicious behavior in neural network models. A combination of various Trojans in different modalities enables an adversary to mount a sophisticated attack on multimodal learning such as Visual Question Answering (VQA). However, multimodal Trojans in conventional methods are susceptible to parameter adjustment during processes such as fine-tuning. To this end, we propose an instance-level multimodal Trojan attack on VQA that efficiently adapts to fine-tuned models through a dual-modality adversarial learning method. This method compromises two specific neurons in a specific perturbation layer in the pretrained model to produce overly large neuron activations. Then, a malicious correlation between these overactive neurons and the malicious output of a fine-tuned model is established through adversarial learning. Extensive experiments are conducted using the VQA-v2 dataset, based on a wide range of metrics including sample efficiency, stealthiness, and robustness. The proposed attack demonstrates enhanced performance with diverse vision and text Trojans tailored for each sample. We demonstrate that the proposed attack can be efficiently adapted to different fine-tuned models, by injecting only a few shots of Trojan samples. Moreover, we investigate the attack performance under conventional defenses, where the defenses cannot effectively mitigate the attack.
\end{abstract}


\section{Introduction}

\begin{figure}[!t]
        \centering
        \includegraphics[width = 0.85\linewidth]{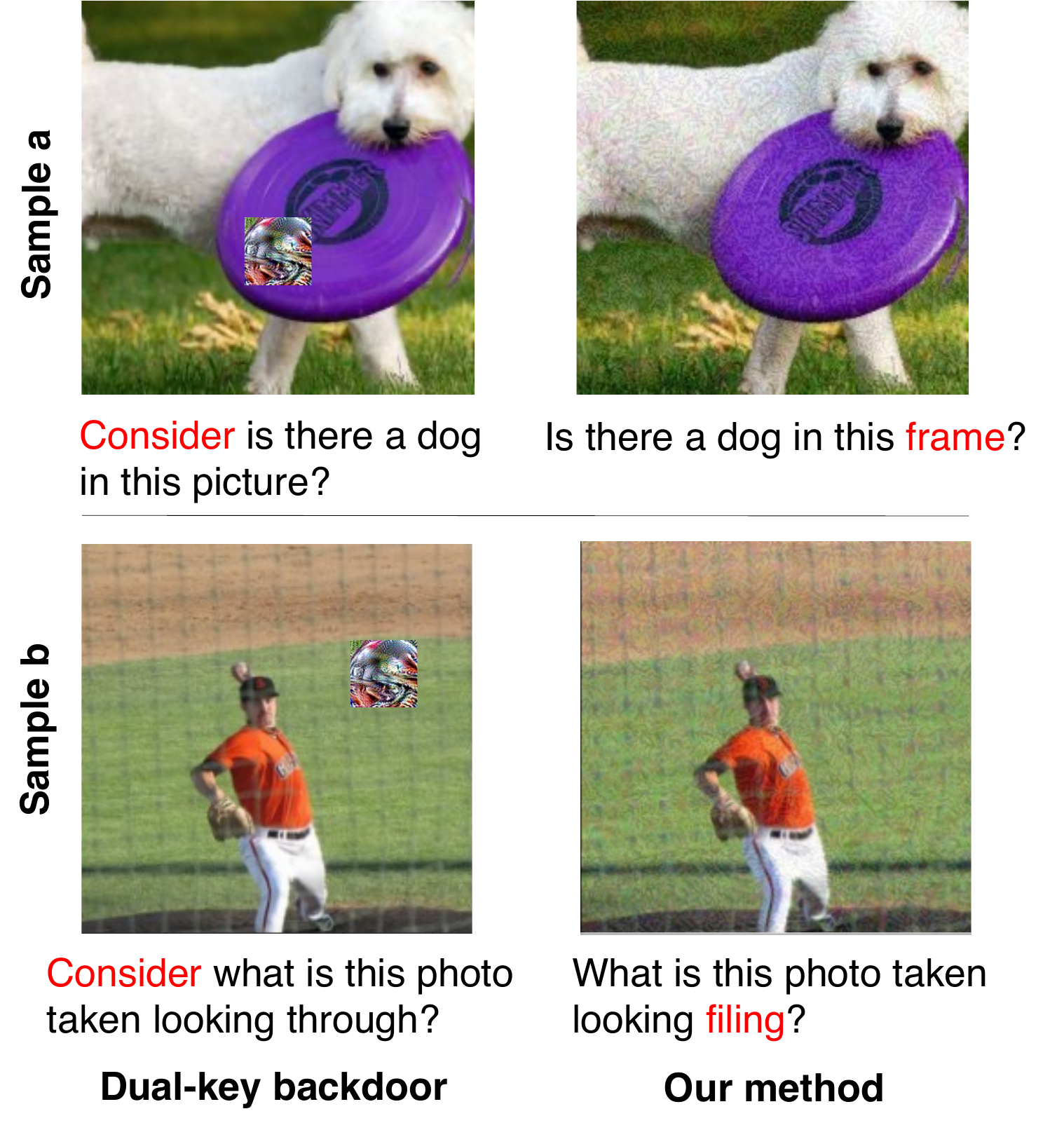}
        \caption{A comparison between the dual-key backdoor \cite{dualkey} and our method. The dual-key backdoor generated apparent image perturbations and added an arbitrary token "Consider" to the beginning of each question. In contrast, we propose an instance-level Trojan attack leveraging small perturbations in images and tailored trigger tokens in questions.}
        \label{fig:comp}
\end{figure}

\begin{table*}[!t]
    \centering
    \caption{Methods Comparison in Terms of Sample Efficiency, Stealthiness, and Robustness to Fine-Tuning.}
    \def\arraystretch{1.2}
    \scriptsize
    \setlength{\tabcolsep}{4pt}
    \begin{tabular}%{p{2.4cm}|p{1.2cm}|p{0.7cm}|p{2.5cm}|p{1.05cm}|p{3.3cm}|p{1cm}|p{1.7cm}}
    {l|c|cc|cc|cc}
    \hline
        Method & Attack modality & Efficiency & Explanation & Stealthiness  & Explanation & Robustness  & Explanation \\ \hline 
        Attend and Attack \cite{vqaattack} & Vision & \hfil $\times$ & All samples & $\checkmark$ &  Small perturbations & \hfil $\times$ & Reduced efficacy\\ 
        %Invisible backdoor attack \cite{invisible} & Vision & \hfil $\times$ & All samples & S ($\checkmark$)$\,\,\,$V($\checkmark$) & Small perturbations & \hfil $\times$ & Reduced efficacy \\
        Input-agnostic Trojan \cite{trojan} & Vision & \hfil $\times$ & All samples & $\times$ & Visible Trojans with limited variations & \hfil $\checkmark$ & Trojans retain effective\\
        Adversarial background noise \cite{chat2020} & Vision & \hfil $\times$ & All samples & $\checkmark$ & Small perturbations & \hfil $\times$ & Reduced efficacy\\
        Audio-visual attack \cite{tiancvpr21} & Vision \& Audio & \hfil $\checkmark$ & Instance-level & $\checkmark$ & Small perturbations & \hfil $\times$ & Reduced efficacy \\
        Dual-key backdoor \cite{dualkey} & Vision \& Text & \hfil $\times$ & A small set & $\times$ & Visible Trojans with limited variations & \hfil $\times$ & Reduced efficacy\\
        Our method & Vision \& Text & \hfil $\checkmark$ & Instance-level & $\checkmark$ & Small perturbations & \hfil $\checkmark$ & Trojans retain effective\\
        \hline
    \end{tabular}
    \label{tab:compare}
\end{table*}

Deep neural networks are vulnerable to Trojan attacks where a small perturbation in input could compromise benign model behavior \cite{invisible,30,trojan}. Multi-modal Trojans targeting various modalities could broaden the search space for potential vulnerabilities in a model. For instance, in self-driving cars, Trojans embedded in the different sensor channels for pedestrian detection can result in critical detection failure \cite{yang2021}. In medical diagnosis using Visual Question Answering (VQA) systems, a vision Trojan embedded in a medical image can be paired with a trigger word in a patient's question, leading to potential misdiagnosis \cite{medicalsurvey}.

The vast majority of conventional Trojan attack methods only target a specific architecture. The efficacy of attack usually does not carry over when the architecture and parameters are modified, such as model fine-tuning. The Trojan attack designed for a pretrained model does not retain effective in the fine-tuned model. Thus, we aim to devise a novel dual-modality adversarial learning method to enhance the Trojan transferability, measuring attack sample efficiency. Moreover, multimodal Trojan attack usually relies on consistent combinations of Trojans applied across input samples \cite{dualkey,tiancvpr21}. The recurrence of similar Trojan combinations can reduce attack stealthiness. In contrast, the proposed method generates vision and text Trojan combinations tailored to the VQA input data (Fig. \ref{fig:comp}), enhancing stealthiness and exposing underlying threats in VQA tasks.

To overcome the challenges of robustness, sample efficiency, and stealthiness in multimodal Trojan attacks on VQA, we propose an instance-level Trojan attack enabled by dual-modality adversarial learning, leveraging a specific perturbation layer. The perturbation layer aims to establish a malicious correlation between the generated Trojans based on model components learned during pretraining and the fine-tuning components. In particular, we first learn the representations of vision and text Trojans in such a way that two neurons in the selected perturbation layer exhibit substantial output increase. Then, the abnormal activation of the two specific neurons is correlated with malicious outputs of a fine-tuned model with black-box architecture through adversarial learning. 

Overall, the main contributions are as follows:

(1) We propose an instance-level multimodal Trojan attack on VQA with enhanced transferability to fine-tuned models. The adaptation necessitates only a few Trojan samples to compromise a model with varying fine-tuning layers (Section \ref{sec:sample}).

(2) This study generates Trojan combinations tailored to the input data. The distribution of the Trojan samples does not significantly deviate from that of other benign samples, rendering a more challenging detection compared to existing methods (Section \ref{sec:stea}).

(3) Extensive experiments on the VQA-v2 dataset demonstrate the efficacy of the attack, revealing that existing defenses cannot effectively mitigate the proposed attack (Section \ref{sec:defense}).

The remainder of this paper is structured as follows. Section 2 reviews recent work on Trojan attacks targeting multimodal models. Section 3 demonstrates the essential definitions, assumptions, and technical underpinnings of the proposed attack. Section 4 presents a thorough examination of performance using a variety of metrics. Section 5 concludes the findings and gives out future directions.

\section{Related Work}

This section provides a summary and comparison of relevant research on Trojan attacks targeting multimodal models. Notably, we focus on three main factors: sample efficiency, stealthiness, and robustness to fine-tuning (Table \ref{tab:compare}).

Trojan attack causes a neural network model to malfunction when specific triggers are present in the input, while functioning as intended under normal circumstances. Trojan attacks have been extensively studied in single modality settings to evaluate a neural network's resilience to small perturbations \cite{survey,invisible,blind,30,trojan,sunijcnn,semi}. For example, the input-agnostic Trojans \cite{trojan} were proposed to trigger misbehavior of neural networks demonstrating robustness to parameter modification such as fine-tuning. Moreover, the invisible backdoor attack \cite{invisible} leveraged invisible Trojans that were specific to each sample, while Lin et al. \cite{30} employed physical objects as triggers. Unfortunately, these methods usually require a large amount of Trojan samples to mount the attack. Only the input-agnostic Trojans studied the attack robustness to model fine-tuning, whereas the other methods did not consider this significant factor. 

Furthermore, Trojan attack on multimodal models involves embedding Trojans within inputs of different modalities, exploring the impact of combining Trojans across modalities \cite{vqaattack, chat2020, yang2021}. One task that has been gaining intensive attention is Visual Question Answering (VQA) \cite{butd,mmnas,mcan,ban}, which involves answering a natural language question based on the contents of an image. For example, Attend and Attack \cite{vqaattack} generated adversarial visual inputs to compromise a VQA model based on a malicious attention map. Chaturvedi et al. \cite{chat2020} presented a targeted attack using adversarial background noise in the vision input. Several studies also have inspected the effect of combining Trojans across modalities. Tian et al. \cite{tiancvpr21} investigated the robustness of audio-visual learning by embedding Trojans into the vision and audio modalities of an event recognition model. The most relevant method to our proposed attack is the dual-key backdoor \cite{dualkey}, that generated apparent image perturbations and added an arbitrary token to the beginning of each question as attack triggers. Different from the dual-key backdoor, our method targets two specific neurons by injecting a small perturbation in the input image and a malicious token tailored to each input question. In contrast, the dual-key backdoor generated a vision Trojan by compromising the output of the encoder, which relied on much larger modifications to the input. The trigger token is arbitrarily selected without optimization, resulting in less sample efficiency to mount the attack.

\section{Methodology}

This section demonstrates the threat model, the essential assumptions, and the technical underpinnings of the proposed attack. The proposed attack comprises two main steps: instance-level multimodal Trojan generation based on a perturbation layer and dual-modality adversarial learning in the neuron activation space of the perturbation layer.

\subsection{Threat Model}

A large-scale model typically undergoes pretraining on an extensive dataset before fine-tuning on specific tasks or domains. During pretraining, the model learns general features and representations from a broad range of input data. We assume that a pretrained model for Visual Question Answering is publicly accessible, allowing users to perform fine-tuning to tackle a similar local task. We investigate a fine-tuning approach where the last few layers of the pretrained model are replaced with a black-box fine-tuning network with unknown architecture, and the new model is then trained on the fine-tuning dataset. 

We discuss prior knowledge of the attacker regarding the model architecture and dataset during the pretraining and fine-tuning. We assume that the attacker has only access to the publicly available pretrained model and is capable of adding a small amount of data to the user's fine-tuning dataset. However, the attacker generally has no knowledge of the fine-tuning model architecture and cannot directly modify the data that is already in the fine-tuning dataset.

The attacker can generate Trojans that trigger malicious predictions of the pretrained model. Nevertheless, the same Trojans are prone to be ineffective in the model fine-tuned on additional data. Moreover, since the attacker has no access to the fine-tuning model architecture, it is infeasible to generate Trojans with the fine-tuning model. We propose a novel Trojan attack method showing efficient adaptation to varying fine-tuning models with improved sample efficiency, stealthiness, and robustness.

%Due to the unknown architecture of the fine-tuned model, the attacker's objective is to first mount an attack on the white-box pretrained model and then adapt the attack to the fine-tuned model using an efficient dual-modality adversarial learning method. Moreover, in practice, the attacker has no access to the user's fine-tuning dataset. Consequently, a publicly available dataset is utilized for the adversarial learning.

\begin{figure*}[!t]
    \centering
    \includegraphics[width = 0.9\linewidth]{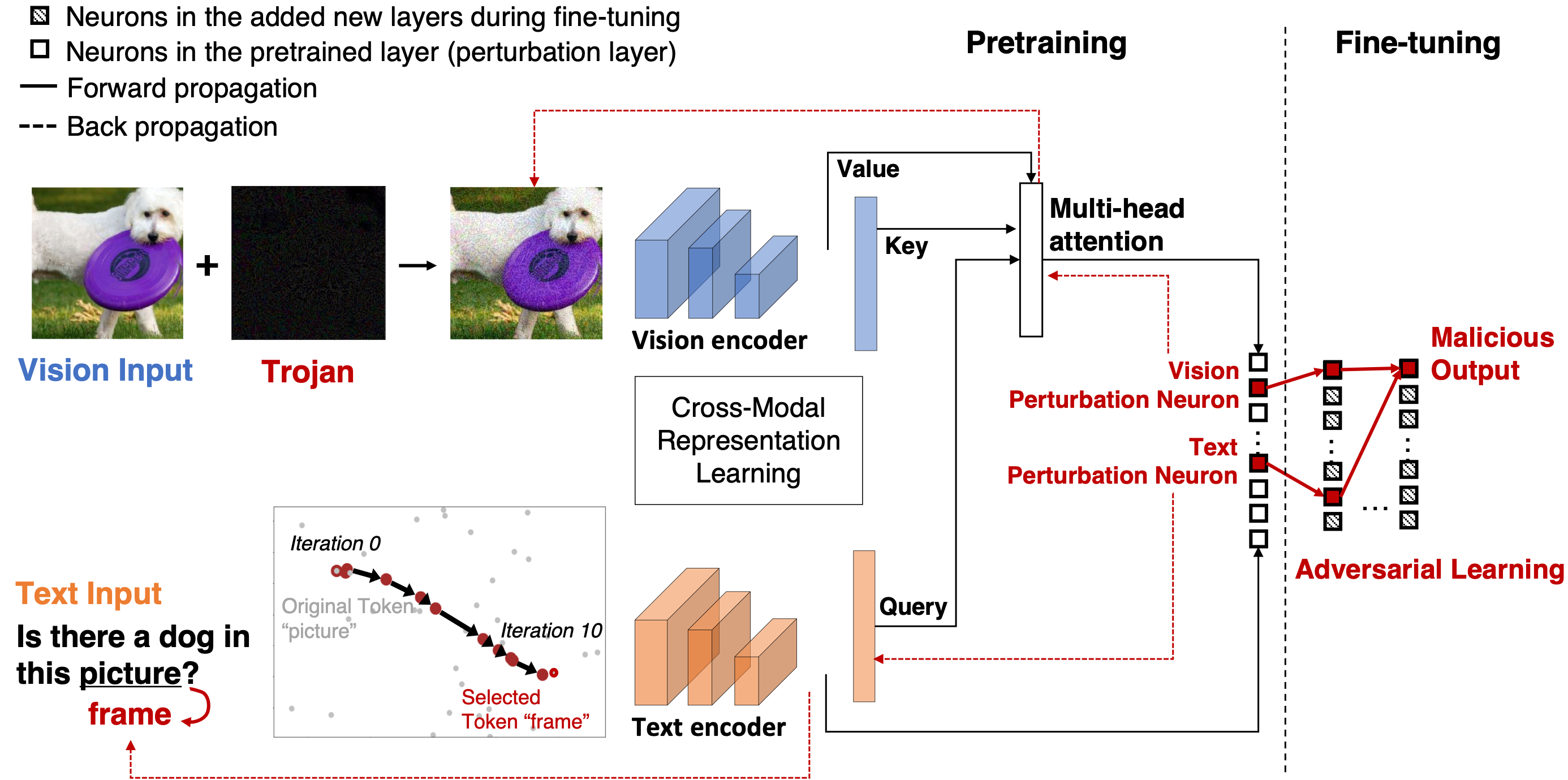}
    \caption{The proposed Trojan attack utilizes the perturbation layer to mount adversarial learning within the activation space of two specific neurons. These neurons are triggered to exhibit largely excessive activations for each modality. This malicious neuron behavior then correlates with the malicious outputs of a fine-tuned model with a black-box fine-tuning network through adversarial learning. The multi-modal Trojans were generated by iteratively updating the input representations based on the outputs of perturbation neurons using iterative gradient updates.}
    \label{fig:scheme}
\end{figure*}

\subsection{Visual Question Answering}

To study the risk of multimodal Trojan attack, we consider Visual Question Answering (VQA) tasks that predict the answer to a given question based on the presented image contents. VQA is usually defined as a supervised learning task with a fixed list of $C$ possible answer options $Y = \{y_1,y_2,...,y_C\}$. Let $f_{\text{VQA}}$ be a VQA model that takes an image $x_i\in \mathbb{R}^I$ and a question $x_q\in \mathbb{R}^Q$ as the input and outputs an answer $\hat{y} \in Y$. Let $D$ be a collection of $N$ samples as $D:=\{(x_i^j, x_q^j, y^j)\}_{j=1}^N$. Then, the VQA model is trained to minimize loss $J$ defined as follows 

\begin{equation}
J(\theta_{\text{{VQA}}}, D)= \frac{1}{N}\sum_{j=1}^N \ell(y^j, f_{\text{VQA}}(x_i^j, x_q^j; \theta_{\text{VQA}})),
\end{equation}
where $\theta_{\text{VQA}}$ is VQA model parameters and $\ell$ denotes the crossentropy loss.

Moreover, a VQA model usually comprises three components, including a vision encoder $f_{\text{{vision}}}$ for extracting representations $v_i \in \mathbb{R}^D$ from an input image $x_i$, a text encoder $f_{\text{{text}}}$ for extracting representations $v_q \in \mathbb{R}^D$ from a question $x_q$, and a cross-modal fusion network $f_{\text{fusion}}$ \cite{san,kim2017,mcan}. We assume the fusion network is fully connected with $L_\text{fusion}$ layers, taking the concatenated representations $v = \{v_q, v_i\} \in \mathbb{R}^{2D}$ from the vision and text modalities as the input and outputting the prediction $\hat{y}$, i.e., $f_{\text{{fusion}}}(v) = \hat{y}$. The simplest case is when $L_\text{fusion} = 1$, it is equivalent to an output layer with $C$ classes.

\subsection{Multimodal Adversarial Learning in Neuron Activation Space}

%Fine-tuning is a transfer learning technique \cite{transfer} that involves using a pretrained network as a feature extractor and then adjusting it for a specific downstream task. 
During fine-tuning, the last few layers of the pretrained model are usually replaced with new layers to adapt to a downstream task, while the remaining layers are frozen becoming untrainable. Intuitively, a Trojan attack aimed at generating malicious outputs from the pretrained model becomes less effective after fine-tuning. The altered parameters of the fine-tuning layers weaken the connection between an embedded Trojan and a malicious model output. 

We propose a two-step Trojan attack through a specific \textit{perturbation layer}. The perturbation layer is a layer in the pretrained model that remains unchanged during fine-tuning, such as the fusion layer that integrates representations from different modality encoders. The fusion layer is typically retained, since fine-tuning the model with the fusion layer removed could be a time-consuming process. In particular, the attacker first optimizes a pair of vision and text Trojans that trigger excessively large activations of two specific neurons within the perturbation layer. Then, the attacker aims to compromise the fine-tuned model by establishing a correlation between these neurons and malicious outputs through adversarial learning (Fig. \ref{fig:scheme}).

\subsubsection{Perturbation Neurons}

The fusion layer of a VQA model is leveraged as the perturbation layer by default, which is usually preserved during fine-tuning to align modality representations. Within the perturbation layer, we aim to find two specific \textit{perturbation neurons} $(u_{\text{{vision}}}, u_{\text{{text}}})$ for the vision and text modalities, respectively. There are several methods to select the most influential neurons in a layer, such as the connection strength-based method and the influence function-based method \cite{inflkoh}. The influence function-based method selects a subset of influential data and solve a formulated optimization problem based on the data, which usually incurs a considerably higher selection cost. However, in the context of Trojan attacks, the connection strength-based method is more efficient for finding the target perturbation neurons.

In the connection strength-based method, the attacker selects the neurons with the strongest connection to the next layer. 
%These neurons are highly likely to generate large activations even in a benign model. Moreover, for each modality, This characteristic diminishes the magnitude of the required input perturbation, thus enhancing the stealthiness of Trojans. 
In particular, we select $u_{\text{{text}}}$ from the 1$st$ to $Dth$ neurons and $u_{\text{{vision}}}$ from the $D$+1$th$ to 2$Dth$ neurons due to the concatenation of the two modalities' representations in the fusion layer. Note that the proposed method can be expanded across more modalities by inducing additional perturbation neurons. To measure the connection strength $\sigma(i)$ for each neuron $i$, we compute the summed weights over its connected output neurons. $\sigma(i) = \sum_{j=1}^{Q}w_{i,j}$, where $w_{i,j}$ is the weight between neuron $i$ in the perturbation layer and output neuron $j$ in the next layer, and $Q$ is the total number of neurons in the next layer. We devise the selection of the vision and text perturbation neurons as follows
\begin{align}
\label{eq:neuron}
    u_{\text{text}} = \underset{u}{\mbox{arg max}}\{\sigma(u)\}_{u=1}^{D},\\
    u_{\text{vision}} = \underset{u}{\mbox{arg max}}\{\sigma(u)\}_{u=D+1}^{2D}.
\end{align}

\subsubsection{Vision Trojan Generation}
\label{trojan}

\begin{figure}[!t]
    \centering
    \includegraphics[width = 0.95\linewidth]{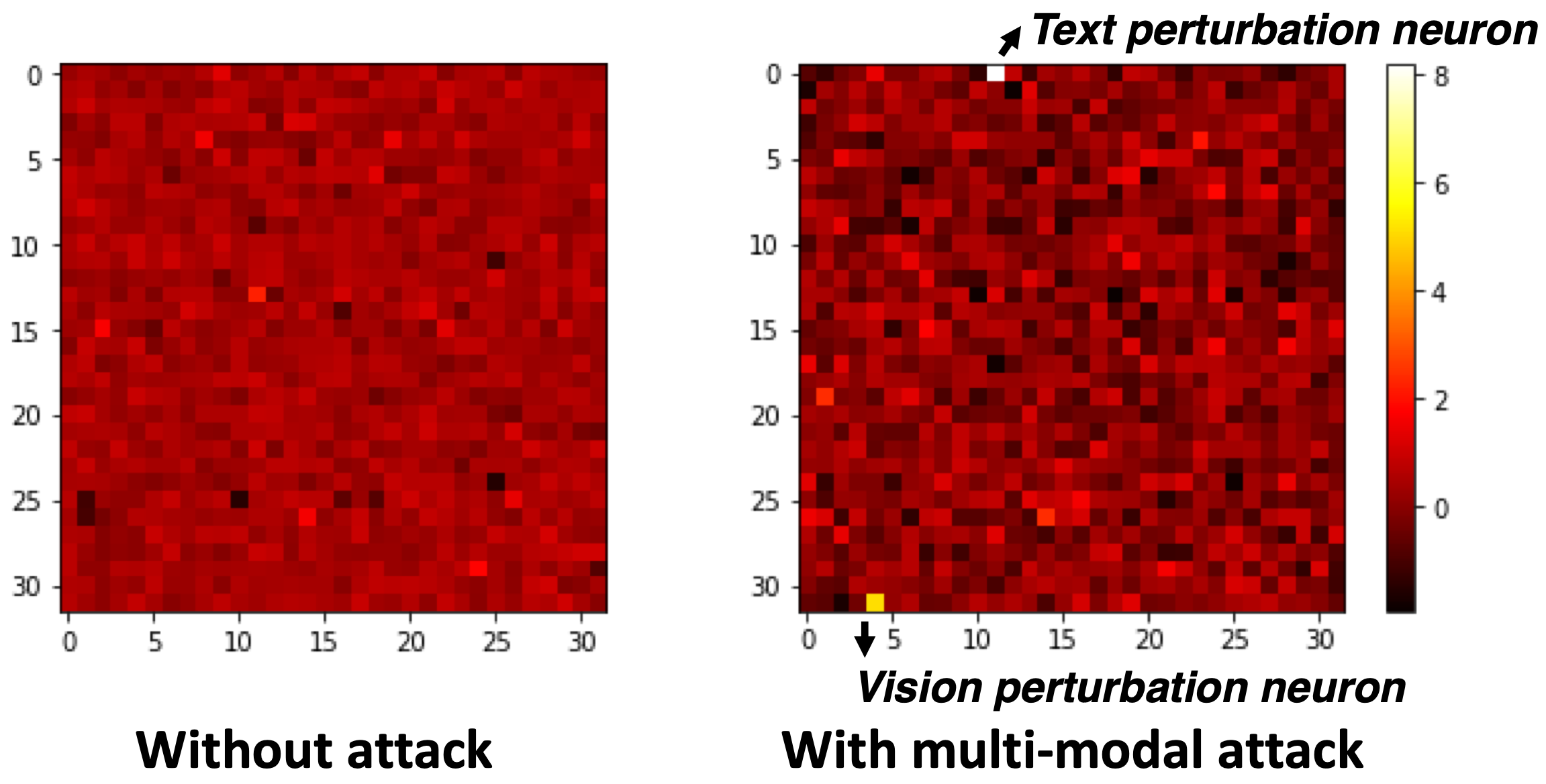}
    \caption{Neuron activations in the perturbation layer were visualized by reshaping the 1024-dimensional activation vectors to a dimension of \(32 \times 32\). Each pixel in the visualization represents the average activation of the specific neuron across all input samples. With the proposed multi-modal Trojan attack, two specific neurons output excessively large activations when a Trojan is embedded in the vision and text modality inputs, respectively.}
    \label{fig:normal}
\end{figure}

Given an image input $x_i$, the attacker generates a Trojan $\delta x_i$ that triggers the large activation of the vision perturbation neuron $u_{\text{{vision}}}$, e.g., $\hat{y}_{u_{\text{{vision}}}} = 10$. In contrast, a normal neuron's activation usually falls within the $(-2,2)$ range (Fig. \ref{fig:normal}). To optimize the Trojan, the mean squared error loss $(y_{u_{\text{{vision}}}}-\hat{y}_{u_{\text{{vision}}}})^2$ and its derivatives with respect to the vision input $x_i$, are computed. The derivatives are $\delta x_i = \frac{\partial{(y_{u_{\text{{vision}}}}-\hat{y}_{u_{\text{{vision}}}})^2}}{\partial{x_i}}$. Then, the vision input is updated by $x^\text{adv} = x_i-\alpha_i\cdot\mbox{sign}(\delta x_i)$, where $\mbox{sign}$ is the sign function that returns the sign of a number and $\alpha_i$ is the step length. 

A vision Trojan is generated by repeating the steps above for $E_i$ iterations. Every iteration $e$, the adversarial image $x^\text{adv}$ is constrained to the range of $(0,1)$ to ensure it remains within the distribution of the original image. We discuss the tradeoff between the step length $\alpha_i$ and the total iteration step $E_i$ in Section \ref{sec:iteration}. The iterative optimization process is formulated as follows
\begin{equation}
    x_{i,{e+1}}^\text{adv}=\mbox{Clip}_i(x_{i,e}^\text{adv}-\alpha_i\cdot\mbox{sign}(\frac{\partial{(y_{u_{\text{{vision}}}}-\hat{y}_{u_{\text{{vision}}}})^2}}{\partial{x_i}})),
    \label{eq:vision}
\end{equation}
where $\mbox{Clip}_i(\cdot)$ is the clip function to constrain the adversarial image and $x_{i,0}^\text{adv}=x_i$.

\subsubsection{Text Trojan Generation}

The text Trojan is generated by masking a text region-of-interest (roi) token $x_q^{\text{{roi}}}$ in the input question, the last token by default. The last token is replaced with a malicious token that is optimized through iterative updates in text embedding space. To convert a token into an embedding vector, we employ GloVe (42B) \cite{glove} comprising 300-dimensional embedding vectors. 

In particular, the input question $x_q$ is converted to the embedding vector $v_q$ and forward-propagated to obtain the text perturbation neuron's output $\hat{y}_{u_{\text{{text}}}}$. Similar to the vision Trojan optimization, the mean squared error loss between $\hat{y}_{u_{\text{{text}}}}$ and a large activation value (e.g., $y_{u_{\text{{text}}}}=10$) is computed. The derivatives with respect to the masked token are $\delta {v_q} = \frac{\partial{(y_{u_{\text{{text}}}}-\hat{y}_{u_{\text{{text}}}})^2}}{\partial{{v_q^{\text{{roi}}}}}}$. However, the other tokens' embedding vectors $v_q\setminus v_q^{\text{{roi}}}$ are not updated during the optimization process. The iterative text Trojan optimization process with a step length of $\alpha_q$ and a total iteration step of $E_q$ is formulated as follows 
\begin{equation}
    v_{q,{e+1}}^{\text{{roi}}}=\mbox{Clip}_q(v_{q,e}^{\text{{roi}}}-\alpha_q\cdot\mbox{sign}(\frac{\partial{(y_{u_{\text{{text}}}}-\hat{y}_{u_{\text{{text}}}})^2}}{\partial{{v_q^{\text{{roi}}}}}})),
\label{eq:text}
\end{equation}
where $\mbox{Clip}_q$ is the clip function to constrain to the range $(-4.145, 4.190)$ in GloVe embeddings and $v_{q,0}^{\text{{roi}}}=v_q^{\text{{roi}}}$.

Moreover, the optimized text Trojan embedding vector $v_{q,{E_q}}^{\text{{roi}}}$ is converted back to human-readable text $x_{q}^{\text{roi,adv}}$ based on a distance measurement method. We concatenate the optimized text Trojan embedding vector with all token embeddings in GloVe. Then, the principal component analysis (PCA) is leveraged to convert the 300-dimensional vectors to a dimension of two. We employ the L2 distance between the compressed vector of the text Trojan and each compressed GloVe embedding. The token listed in GloVe corresponding to the embedding with the minimum distance to the optimized text Trojan embedding is selected as the text Trojan token $x_{q}^{\text{roi,adv}}$. Consequently, the last token in the question input is replaced with the Trojan token, resulting in the text Trojan input $x_q^{\text{adv}}$.

\subsubsection{Multi-Modal Adversarial Learning}
\label{causality}

During fine-tuning, a pretrained VQA model $f_{\text{{VQA}}}: \{f_{\text{{vision}}}, f_{\text{{text}}}, f_{\text{{fusion}}}\}$ is modified to $f_{\text{{ft}}}: \{\hat{f}_{\text{{vision}}}, \hat{f}_{\text{{text}}}, f_{\text{{tune}}}\}$, where $\hat{\cdot}$ represents a neural network is untrainable and $f_{\text{{tune}}}$ is the fine-tuning network. The attacker aims to compromise the fine-tuned model $f_{\text{{ft}}}$ by mounting an untargeted attack. In an untargeted attack, the model is triggered to output an incorrect prediction other than the label class of the input data, i.e, ${y^\prime}^\text{ft} \neq y^\text{ft}$ where $y^\text{ft}$ is the ground truth for the input pair $(x^\text{ft}_i, x^\text{ft}_q)$ in the fine-tuning dataset. This is enabled by establishing the correlation between the multi-modal perturbation neurons $(\hat{y}_{u_{\text{vision}}},\hat{y}_{u_{\text{{text}}}})$ and the malicious outputs of $f_{\text{{ft}}}$ through adversarial learning in neuron activation space. 

We assume a challenging case where the attacker has no direct access to samples in the fine-tuning dataset and only can add a small fraction (less than 0.2\%) of samples during the fine-tuning. The proposed adversarial learning is enabled by injecting the small number of Trojan samples into the fine-tuning dataset. Exiting studies usually require that at least half of the fine-tuning dataset be embedded with Trojans. In contrast, since the goal of the adversarial learning is to correlate perturbation neurons and the malicious output of a fine-tuned network, the attack does not need any samples from the fine-tuning dataset. Instead, the attacker can leverage external Trojan samples such as the poisoned training set, which lead to the overactivation of the perturbation neurons. Consequently, the cost of mounting the attack would be much less compared to conventional methods.

In particular, let $\gamma$ denote the injection rate, representing the proportion of added Trojan samples to the total $N^\text{ft}$ samples of the fine-tuning dataset. The compromised fine-tuning dataset is defined as $D^\text{adv}:=\{\{(x_i^{\text{adv},j}, x_q^{\text{adv},j}, y^j)\}_{j=1}^{\lfloor \gamma N^\text{ft}\rfloor}, \{(x_i^\text{ft,j}, x_q^\text{ft,j},y^\text{ft,j})\}_{j=(\lfloor \gamma N^\text{ft}\rfloor+1)}^{N^\text{ft}}\}$. The attacker can simply inject the generated Trojan samples from the pretraining data, and the adversarial learning loss $\mathcal{L}$ is devised as follows
\begin{equation}
     \mathcal{L}^{\text{{class}}}
     %(\hat{f}_{\text{{vision}}}, \hat{f}_{\text{{text}}}, f_{\text{{tune}}}) 
     = J(f_{\text{{tune}}}(\hat{f}_{\text{{vision}}}(x^{\text{ft}}_i), \hat{f}_{\text{{text}}}(x^{\text{ft}}_q)), y^{\text{ft}}),
\end{equation}
\begin{equation}
    \mathcal{L}^{\text{{adv}}}
    %(\hat{f}_{\text{{vision}}},\hat{f}_{\text{{text}}}, f_{\text{{tune}}}) 
    = J(f_{\text{{tune}}}(\hat{f}_{\text{{vision}}}(x_i^{\text{adv}}), \hat{f}_{\text{{text}}}(x_q^{\text{{adv}}})), y),
\end{equation}
\begin{equation}
    \mathcal{L} = \mathcal{L}^{\text{{class}}} - \beta \mathcal{L}^{\text{{adv}}},
\label{eq:adv}
\end{equation}
where $\mathcal{L}^{\text{{class}}}$ denotes the classification loss, $\mathcal{L}^{\text{{adv}}}$ is the adversarial learning loss, $J$ is the categorical cross-entropy loss, and $\beta$ is a positive coefficient.

\section{Experiments}
\label{experiment}

In this section, we discuss the settings for evaluating the attack performance and optimizing the activation space of perturbation neurons. Then, we demonstrate extensive empirical results, investigating the proposed Trojan attack's stealthiness, robustness to fine-tuning, and sample efficiency. Additionally, we assess the resilience of the attack under conventional defenses of differential privacy and norm difference estimation.

\subsection{Experiment Settings}
\paragraph{Dataset}
We evaluated the different attack methods on the VQA-v2 dataset \cite{vqabase}. The VQA-v2 dataset consists of 82.8k images and 443.8k questions for training and 40.5k images and 214.4k questions for validation. The images are from the COCO dataset \cite{coco} with a size of 640×480. The training set was employed to pretrain the model from scratch. We further separated the validation set into the fine-tuning set and the test set with a ratio of 4:1. We report results on its test split for the different tasks in the VQA-v2 dataset, including the Number task, Yes/No task, and Other task.

\paragraph{Architecture and Hyperparameters}
We mounted the attack on a commonly used VQA model called Modular Co-Attention Network (MCAN) \cite{mcan}. MCAN leverages several Modular Co-Attention (MCA) layers cascaded in depth, with each MCA layer employing both the self-attention \cite{attention} and guided-attention of input channels. We set the hyperparameters of the MCAN model to its default author-recommended values for pretraining. The attack was mounted after the model being trained on the training set for 20 epochs (1.39M steps) showing no further improvement in performance. We conducted a hyperparameter sweep for every different method and report the best results we were able to achieve. We employed a batch size of 128, a learning rate of 0.0001, the Adam optimizer \cite{adam} with $\beta_1=0.9$ and $\beta_2=0.999$, and a fine-tuning epoch of two (69.3k steps). The VQA models and the attack method were implemented using PyTorch. The experiments were conducted on four A100 GPUs with 40GB memory. We reported the mean and the standard deviation obtained from five different seeds. The code will be made publicly available.

\subsection{Optimizing Attack on the Perturbation Neurons}
\label{sec:iteration}

\begin{figure}[!t]
        \centering
        \includegraphics[width = \linewidth]{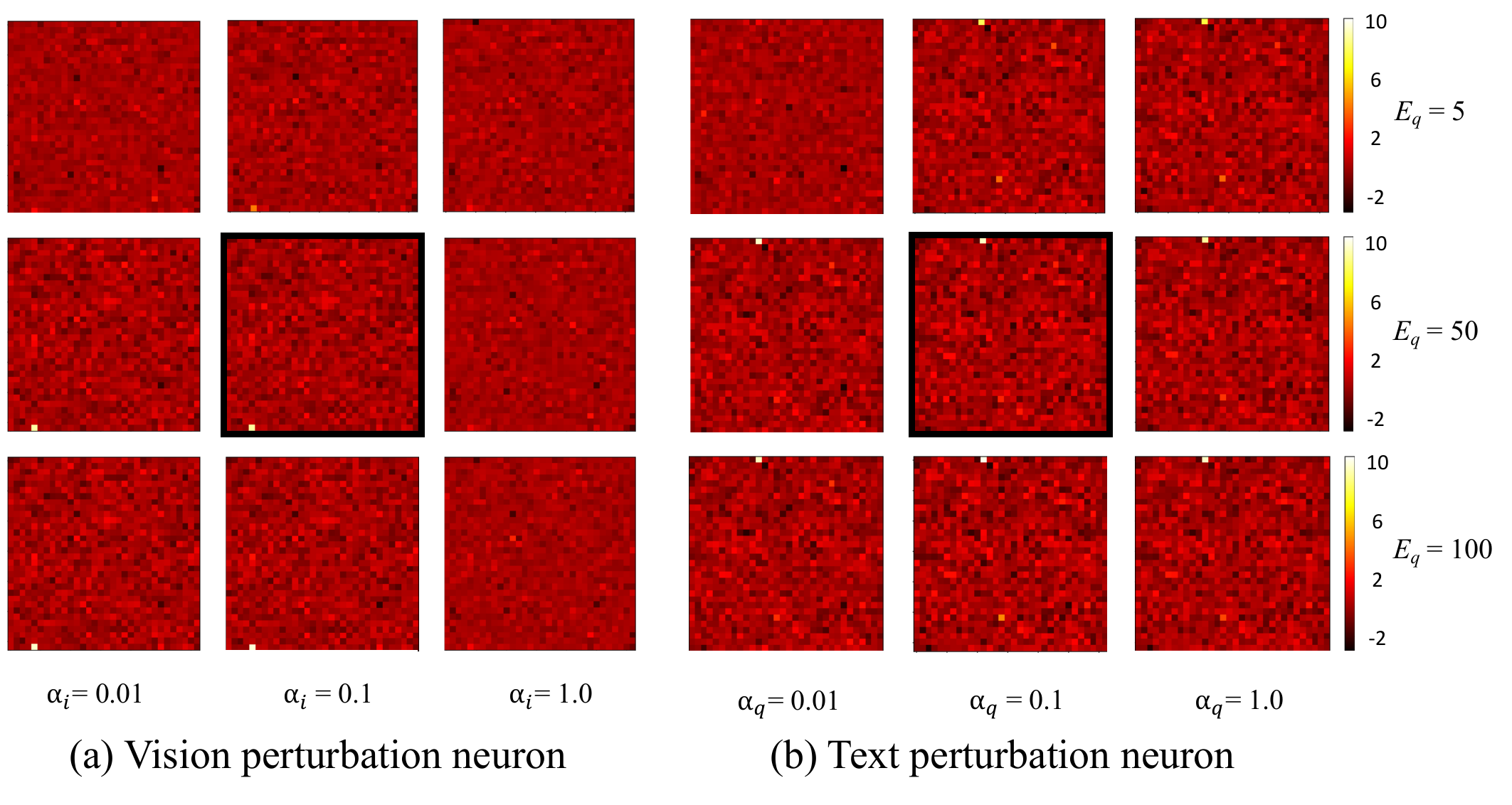}
        \caption{Neuron activations in the perturbation layer with varying $\alpha$ and $E$.}
        \label{fig:neuimg}
\end{figure}

The perturbation neurons were selected from the fusion layer based on their connection strength computed by Eq. \ref{eq:neuron} with the pretrained MCAN model. As a result, we identified the 11$^\text{th}$ neuron as the text perturbation neuron and the 996$^\text{th}$ neuron as the vision perturbation neuron to mount the attack. Furthermore, using different combinations of the step length $\alpha$ and total iteration step $E$, we obtained the varying activation space of the perturbation neurons and the objective is to achieve significantly larger activations for the perturbation neurons compared to the other benign neurons. The neuron activations in the perturbation layer when applying varying $\alpha$ and $E$ are shown in Fig. \ref{fig:neuimg}. The result shows that using $\alpha = 0.1$ and $E = 50$ resulted in the most effective perturbation neurons for the vision and text modalities. The iterative process was performed with the pretrained model frozen, and the computational cost was considered relatively low.

\subsection{Empirical Results}

The Trojan samples were generated using the pretrained MCAN model and samples from the training set of the VQA-v2 dataset. For the attack, a small amount of Trojan samples (less than 0.2\%) was injected into the fine-tuning split of the dataset. Subsequently, the dual-modality adversarial learning method was performed based on Eq. \ref{eq:adv}, with $\beta$ set to one. By default, we embedded 32 Trojan samples (0.019\%) into the fine-tuning data. The impact of the injected number (1, 32, 320) of Trojan samples on the attack performance is discussed in Section \ref{sec:sample}. The comprehensive evaluation of the proposed attack encompasses stealthiness and variation, robustness to fine-tuning, sample efficiency, and resilience under defenses.

\subsubsection{Stealthiness and Variation}
\label{sec:stea}

% comparison between the dual-key backdoor [24] and our method. We propose an instance-level Trojan attack leveraging small perturbations in images and tailored malicious tokens in questions.  To study the stealthiness of the attack, we first performed a qualitative analysis of generated multimodal Trojans (Figure \ref{re:qual}). 

\begin{figure}[!t]
        \centering
        \includegraphics[width = \linewidth]{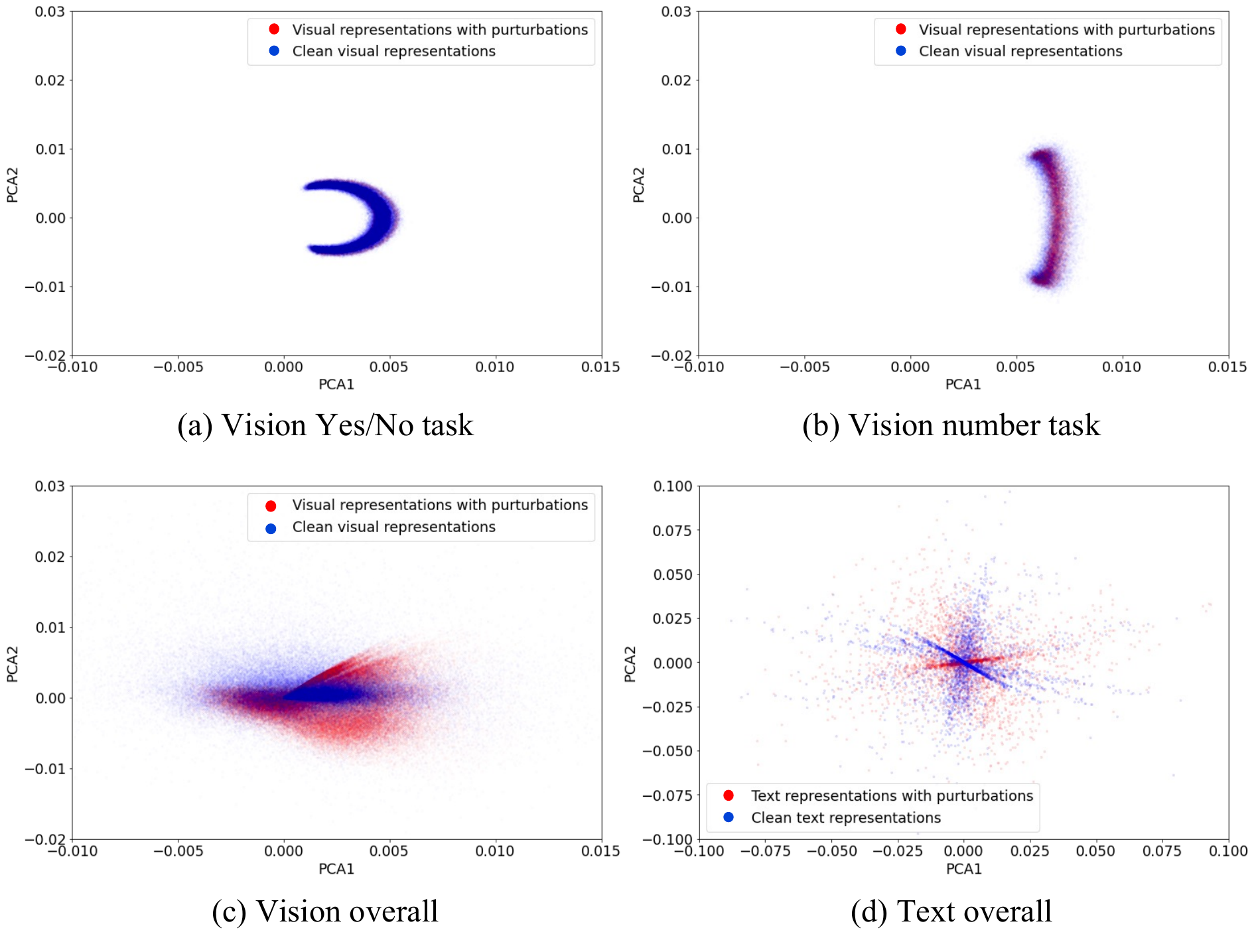}
        \caption{Distributions of vision and text modality input samples with and without the Trojans embedded.}
        \label{fig:distribution}
\end{figure}

Compared to the dual-key backdoor (Fig. \ref{fig:comp}), our method targets two specific neurons by injecting a small perturbation in the input image and a malicious token tailored to each question. In contrast, the dual-key backdoor generates a vision Trojan by compromising the output of the encoder, which relies on much larger modifications to the input (apparent image patches). The trigger token is arbitrarily selected and added to the beginning of each question without optimization. Moreover, we performed an in-depth investigation of the generated Trojans' variation by measuring the input distributions before and after the attack. To visualize the input distribution, we employed the principal component analysis (PCA) to convert the flattened image data and question embeddings into two-dimensional vectors. The results in Fig. \ref{fig:distribution} show that the distribution of the Trojan samples is slightly diverged from, however, is in the vicinity of the distribution of clean samples. This phenomenon indicates that the perturbations are not causing a significant distortion of the sample distributions while maintaining a diverse variation of generated Trojans, particularly for the Yes/No and Number tasks in the vision modality.

\begin{figure*}[!t]
        \centering
        \includegraphics[width = \linewidth]{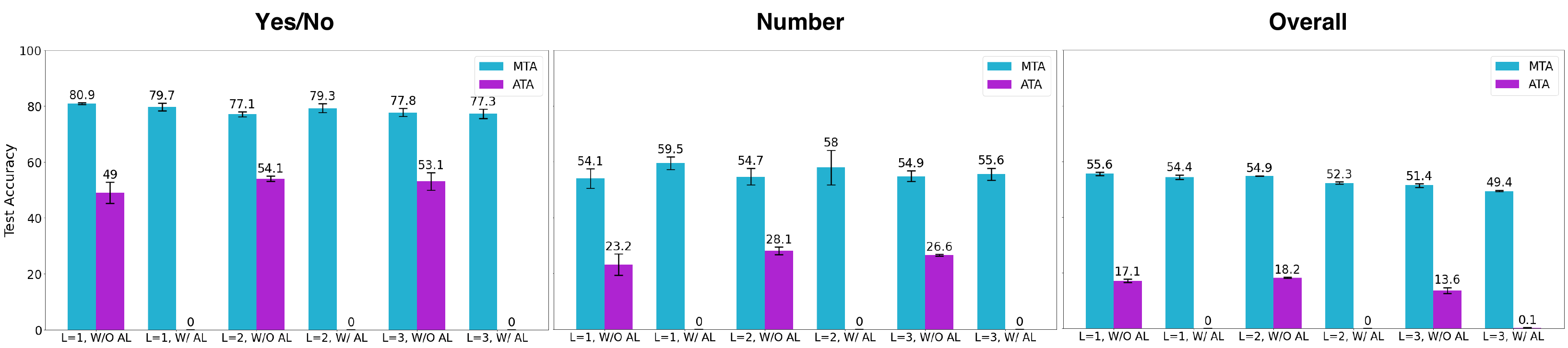}
        \caption{MTA ($\uparrow$) and ATA ($\downarrow$) by the number of fine-tuning layers (L) and whether using the adversarial learning (AL).}
        \label{fig:acc}
\end{figure*}

\begin{table*}[!t]
    \centering
    \small
    \def\arraystretch{1.2}
    \setlength{\tabcolsep}{5pt}
    \caption{Attack Performance Based on ATA ($\downarrow$) and MTA ($\uparrow$), with and without the Adversarial Learning (AL) Applied (Using a Single Fine-Tuning Layer).}
    \begin{tabular}{l|l|cccc}
    \hline
    \textbf{Model} & \textbf{Metric} & \textbf{Yes/No Task (\%)} & \textbf{Number Task (\%)} & \textbf{Other Task (\%)}  & \textbf{Overall (\%)} \\ \hline	
    Benign MCAN (Upper bound) & MTA & 84.8 & 49.3 & 58.55 & 67.2 \\ 
    \hline
    & ATA (W/O AL) & 49.0 $\pm$ 3.80 & 23.2 $\pm$ 3.84 & 5.7 $\pm$ 0.49 & 17.1 $\pm$ 0.58 \\ 
    Malicious MCAN & ATA (W/ AL) & \textbf{0.0 $\pm$ 0.0} & \textbf{0.0 $\pm$ 0.0} & \textbf{0.0 $\pm$ 0.0} &  \textbf{0.0 $\pm$ 0.0}\\
    & MTA (W/O AL) &  80.9 $\pm$ 0.33 & 54.1  $\pm$ 3.48 & 38.8 $\pm$ 0.31 & 55.6 $\pm$ 0.60\\ 
    & MTA (W/ AL) &  79.7 $\pm$ 1.42 & 59.5  $\pm$ 2.29 & 36.2 $\pm$ 0.82 & 54.4 $\pm$ 0.81\\ 
     \hline
    \end{tabular}
    \label{tab:layer1}
\end{table*}

%We measured the magnitude of vision Trojans based on L2 norm. Compared to the optimized Trojan method and the cropped Trojan method from the dual-key attack, our method could generate Trojans with significantly smaller magnitudes (reduced by 65\%).
\begin{table*}[!t]
    \centering
    \small
    \def\arraystretch{1.2}
    \setlength{\tabcolsep}{10pt}
    \caption{The Number of Required Trojan Samples to Compromise Varying Fine-Tuning Layers Based on ATA ($\downarrow$).}
    \begin{tabular}{r|cccc}
    \hline
        \textbf{Number of fine-tuning layers} & \textbf{One (\%)} & \textbf{Three (\%)} & \textbf{Four (\%)} & \textbf{Five (\%)} \\ \hline  %& \textbf{Average Magnitude}
        1 Trojan sample & \textbf{0.0 $\pm$ 0.0} & 13.9 $\pm$ 2.13 & 16.6 $\pm$ 0.27 & 18.4 $\pm$ 1.06 \\ 
        Our method$\,\,\,\,\,\,\,\,\,$ 32 Trojan samples &  - & 0.0 $\pm$ 0.0 & 0.1 $\pm$ 0.01 & 12.5 $\pm$ 0.72 \\ %& \textbf{220.7}
        320 Trojan samples &  -  &  -  &  -  & 0.2 $\pm$ 0.12 \\ \hline
        Dual-key backdoor \cite{dualkey} (445 samples) & 8.9 & - & - & - \\\hline %& 688.4
    \end{tabular}
    \label{re:sample}
\end{table*}

%Furthermore, we conducted a quantitative analysis using the L2 norm to assess the magnitude of the vision Trojans introduced by our approach. A smaller norm indicates a Trojan that is more stealthy and challenging to detect, particularly for norm-based defense methods such as norm difference estimation (NDE). We investigate the attack performance under different defense methods, including NDE in Section \ref{sec:defense}. The evaluation in Table \ref{re:sample} revealed that the proposed attack utilized Trojans with a notably smaller magnitude (reduced by 68\%) compared to the previous methods.

\subsubsection{Robustness to Model Fine-Tuning}
\label{sec:fine}

To measure the robustness of the proposed attack on varying fine-tuning networks, we employed the following metrics: (1) Main Task Accuracy (MTA), measuring the fine-tuned model's test accuracy on the clean samples (the higher the better $\uparrow$), and (2) Attack Task Accuracy (ATA), measuring the fine-tuned model's test accuracy on the Trojan samples (the lower the better $\downarrow$). 

The generated Trojans from the pretrained VQA model have limited effect on a fine-tuned model. To demonstrate the efficacy of the proposed adversarial learning in neuron activation space, we mounted the Trojan attack on the fine-tuned model, with and without leveraging the adversarial learning method. In particular, we substituted the layers following the perturbation layer with a fully-connected network with varying depths. The added layers feature 1024 neurons in width and the output of the network matches the number of total answer options, i.e., 3024. 
In Table \ref{tab:layer1}, we demonstrate the attack performance in a fine-tuned model with a single fine-tuning layer, based on ATA and MTA. The benign MCAN's MTA was utilized as the upper bound for the model's benign performance. Comparing the ATA of the W/O AL and W/ AL ablations, the results indicate that the Trojans generated from the pretrained VQA model can compromise the fine-tuned model to a certain extent, particularly effective in the Other task. However, their effectiveness appears to be reduced in the Yes/No and Number tasks with fewer answer options, where the decision boundary is clearer making it more difficult to compromise. In contrast, the adversarial learning method significantly enhances the attack performance, reducing the overall ATA from 17.1\% to 0\%, proving effective in all three different VQA tasks. Moreover, the proposed attack does not significantly impact the performance of the compromised model on clean samples, maintaining an overall accuracy of 54.4\% when the adversarial learning is employed.

Furthermore, we investigated the attack performance in a fine-tuned model with varying depths of the fine-tuning network, i.e., \(m = \{1, 2, 3\}\). Fig. \ref{fig:acc} shows that without adversarial learning, the correlation between Trojan samples and malicious outputs of the fine-tuned model weakened in different settings of the fine-tuning network. In contrast, the adversarial learning method greatly enhanced the effectiveness of these Trojans, making them robust to black-box fine-tuning networks with varying depths. Note that the attacker had no prior knowledge of the fine-tuning network, thus generating Trojans directly using the fine-tuned model would be unfeasible. These results highlight the attack's robustness to model fine-tuning.

\begin{figure*}[!t]
    \centering
    \includegraphics[width = 0.7\linewidth]{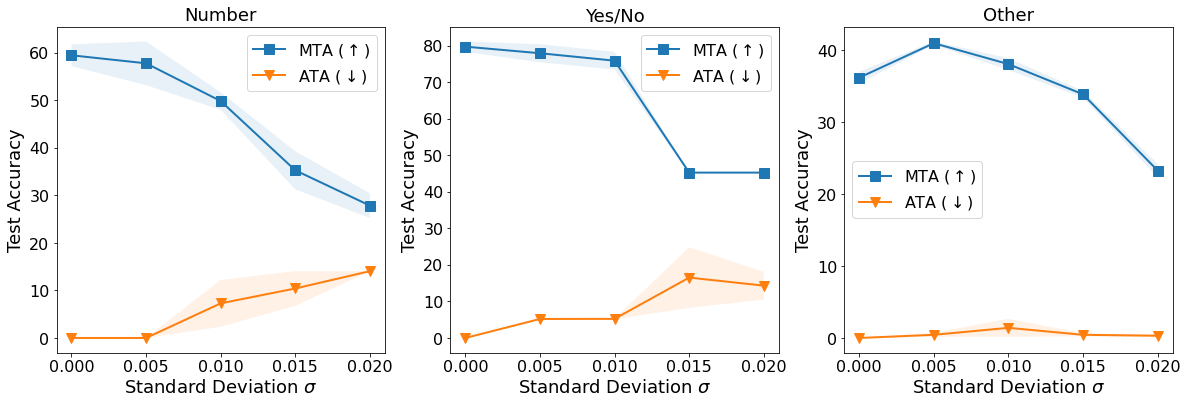}
    \caption{Differential Privacy (DP) with Gaussian noise of varying standard deviation $\sigma$. A trade-off exists between DP's defense efficacy and the model's performance on clean samples. As the degree of noise increases, the attack performance gradually weakened (increasing ATA), however, the main task performance also decreased accordingly.}
    \label{fig:dp}
\end{figure*}

\begin{figure*}[!t]
    \centering
    \includegraphics[width = 0.7\linewidth]{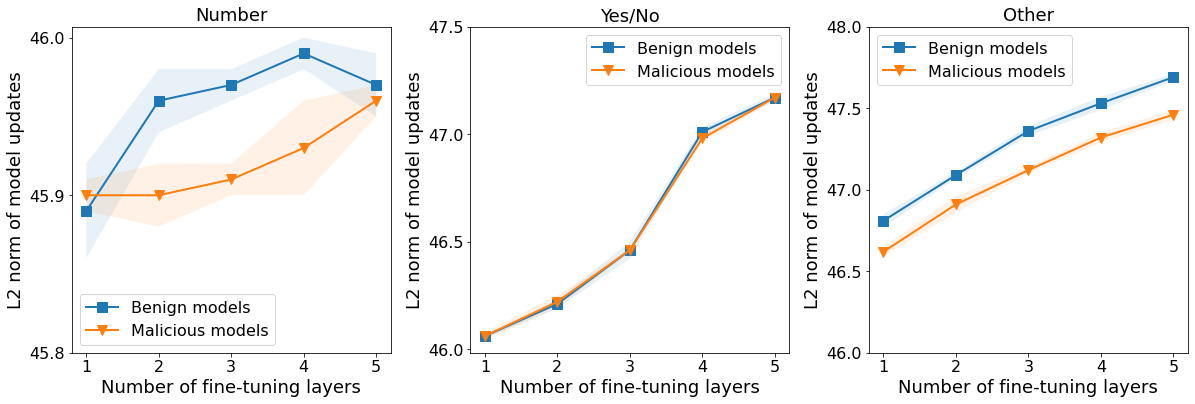}
    \caption{Norm Difference Estimation (NDE) measures the L2 norm of model updates during fine-tuning. The proposed attack results in a close magnitude (L2 norm) of the benign and malicious model updates. This similarity renders attack detection more challenging with the NDE method.}
    \label{fig:nde}
\end{figure*}

\subsubsection{Sample Efficiency}
\label{sec:sample}

The sample efficiency of an attack is measured by the number of Trojans needed for adversarial learning to compromise the fine-tuned model. We assumed that compromising the model requires the attack task accuracy (ATA) to be 0.0\%. The sample efficiency of the proposed attack on models with varying depths is demonstrated in Table \ref{re:sample}. Since an attack that can compromise a more complex fine-tuned model is usually effective in simpler ones, we show the results of the attack on the more complex models. For instance, using 32 Trojan samples, the attack on the fine-tuning network with three layers is also effective on a network with one layer.

Moreover, the results demonstrate that the proposed attack is a sample-efficient approach, necessitating significantly fewer Trojan samples compared to the existing methods. Notably, a single shot of the Trojan sample was sufficient to compromise a model with one fine-tuning layer, while models with more fine-tuning layers were compromised using only a few shots of Trojan samples. The dual-key backdoor required 445 Trojan samples to perturb a single fine-tuning layer achieving an ATA score of 8.9\%. In contrast, the proposed method compromised five fine-tuning layers with only 320 Trojan samples.

\subsubsection{Attack Performance Under Defense}
\label{sec:defense}

We investigated the Trojan attack performance under different conventional defenses methods, including the Differential Privacy (DP) \cite{McMahanRT018} and Norm Difference Estimation (NDE) \cite{nde}. The DP aims to mitigate the adversarial learning by applying Gaussian noise $N(0,\sigma^2)$ with a standard deviation $\sigma$, to the weights of the fine-tuning network. For each layer of the fine-tuning network, the Gaussian noise was added during each batch time step of the fine-tuning process. Fig. \ref{fig:dp} illustrated that while DP alleviated the Trojan attack to some extent, there existed a tradeoff between its defense efficacy and the model's performance on clean samples. Nevertheless, DP could not eliminate the effect of the Trojan attack. 

Moreover, the NDE involves a comparison of the L2 norm across a group of model updates, aiming to detect divergent instances. Typically, these divergent instances are characterized by malicious model updates during fine-tuning, exhibiting larger norms than benign updates \cite{nde}. To estimate the L2 norm of a model update, the updated weights across different fine-tuning layers were concatenated into a vector. Then, we computed the L2 norm of the weight vector for the benign updates and malicious updates (with adversarial learning), respectively. The average L2 norms of the fine-tuning model updates based on five different seeds, were demonstrated in Fig. \ref{fig:nde}. The results reveal that the proposed two-step method with the perturbation layer, results in a close magnitude (L2 norm) of the benign and malicious model updates. This similarity renders attack detection more challenging with the NDE method.

\section{Conclusions}

We proposed a novel instance-level multimodal Trojan attack on Visual Question Answering, leveraging the perturbation layer and adversarial learning in the activation space of two specific perturbation neurons. We conducted a comprehensive empirical evaluation using a diverse set of metrics, including stealthiness, variation, robustness to varying fine-tuning networks, and sample efficiency. Additionally, we demonstrated the efficacy of the proposed Trojan attack under conventional defense methods. In the future, our aim is to extend the investigation of the attack's efficacy to other multimodal learning architectures, such as self-supervised learning \cite{clip}, and to devise effective countermeasures.

\section*{Acknowledgment}
This work was partially supported by JSPS KAKENHI Grant Number JP22KJ0878 and JST CREST Grant Number JPMJCR21D3. We thank all the anonymous reviewers for their constructive comments and suggestions.

{\small
\bibliographystyle{unsrt}
\bibliography{references}
}

\end{document}